# Deconstructing Word Embedding Models

Koushik K. Varma

Ashoka University, koushik.varma_ug18@ashoka.edu.in

*Abstract* – Analysis of Word Embedding Models through a deconstructive approach reveals their several shortcomings and inconsistencies. These include instability of the vector representations, a distorted analogical reasoning, geometric incompatibility with linguistic features, and the inconsistencies in the corpus data. A new theoretical embedding model, 'Derridian Embedding,' is proposed in this paper. Contemporary embedding models are evaluated qualitatively in terms of how adequate they are in relation to the capabilities of a Derridian Embedding.

## 1. INTRODUCTION

Saussure, in the *Course in General Linguistics,* argued that there is no substance in language and that all language consists of differences. In language there are only forms, not substances, and by that he meant that all apparently substantive units of language are generated by other things that lie outside them, but these external characteristics are actually internal to their make-up. All elements of language have identity only in so much as they are produced by a network of differences and therefore, the value of these elements is established by its relation to other elements. (Saussure 1966) This relation can be thought paradigmatically, in terms of substitution, or syntagmatically, in terms of concatenation as the following:

*The value of a sign is given by the positions it can occupy, in opposition to the ones it cannot.*

Or as JR Firth plainly puts it, "*You shall know a word by the company it keeps.*" (Firth, 1957, p. 11) Quantifying the works of these linguists is the domain Distributional Semantics which is based on the hypothesis that words with similar meanings occur in similar contexts (Harris, 1954)

Built on top of this hypothesis are Word Embeddings (Mikolov et al., 2013): unsupervised techniques used to map words or phrases from a text to a corresponding vector of real numbers. The obtained vector space through embedding models preserve the contextual similarity of words – therefore words that appear regularly together in text will also appear together in the vector space.

With embeddings playing a crucial role in downstream NLP tasks such as: Word Sense Disambiguation, POS – Tagging, Named Entity Recognition, Semantic Dependency Parsing and Sentiment Analysis, one must be certain that all linguistic aspects of a word are captured within the vector representations because any discrepancies could be further amplified in practical applications. While these vector representations have widespread use in modern natural language processing, it is unclear as to what degree they accurately encode the essence of language in its structural, contextual, cultural and ethical assimilation. Surface evaluations on training-test data provide efficiency measures for a specific implementation but do not explicitly give insights on the factors causing the existing efficiency (or the lack there of).

We henceforth present a deconstructive review of contemporary word embeddings using available literature to clarify certain misconceptions and inconsistencies concerning them.

### 1.1 DECONSTRUCTIVE APPROACH

Derrida, a post-structuralist French philosopher, describes the process of spatial and temporal movement that he claims makes all thought and reality possible. He set about demonstrating that ideas especially are like units of language; they are generated by difference; they have no substance apart from the networks of differences (each bearing the traces of other elements and other differences) that generate them as effects. (Derrida 1982)

Drawing upon his work, I define the theoretical existence of a *'Derridian Embedding'* that transposes his network of differences into a vector-space model.

The following would be the properties of such an embedding framework:

- All lexical relations of language (Hyponymy, homonymy, polysemy, synonymy, antonymy and metonymy) ought to be captured within the vector representations.

- The resultant vectors should provide a linear mapping between concepts in cognitive thought and elements of language. This presupposes that even the most abstract concepts (such as *love, success, freedom, good,* and *moral*) represented in a vector space are analogous to neural (of the human brain) representations of the concept.

- All possible discourse structures ought to be accommodated as traversals within the network of differences.

I consider Derridian Embeddings to be the epitome of natural language understanding. I then tackle contemporary word embeddings at various levels to be striving towards the end of reaching the capabilities of a *Derridean Embedding* and point out reasons as to why they fail.

In particular, my approach in this paper is the following:

1. Isolate a constraint that limits proper functioning of a word embedding.
2. Analyze the repercussions of the constraint and potential solutions to it.
3. Assume that the constraint has been dissolved and move to another higher-level constraint that exists within the framework.
4. Go to *2*.

For instance, *Section: 3* discusses the instability in embedding models. Further analysis of available literature shows that, the discontinuity in domain-based senses is a major reason for instability. One of the solutions to this is to use an ensemble of domain-based models. Although this isn't an absolute solution, I make an assumption that the instability caused by domain-based senses to be nullified. We then verify if there could be any incompatibility within the embeddings models after the above assumption has been made.

At each constraint, we end up looking at word embeddings through a lens broader than that of the previous constraint.

*Fig 1.0* illustrates different constraints on word embeddings as a part of the deconstruction approach that will be presented over a series of sections in the rest of the paper. The higher in the hierarchy a constraint is, the more broader and complex its repercussions are.

## 2. EMBEDDING MODELS

Word embeddings (Mikolov et al., 2013) are scalable, unsupervised, contemporary models of distributional semantics with the primary objective being the mapping of linguistic terms into real-valued vectors. The obtained vector space preserves the contextual similarity of words – therefore words that appear in similar contexts tend to appear in close proximity in the vector space.

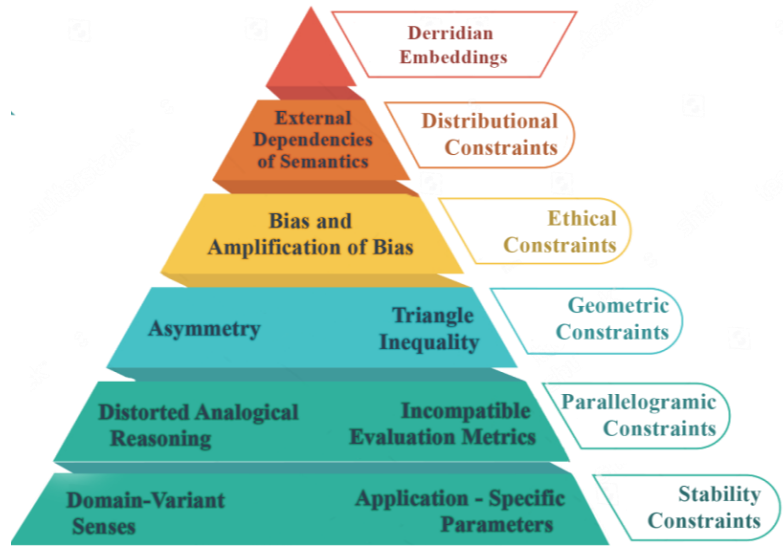

*Fig. 1.0 Overview of constraints to be discussed in this paper (Bottom to Top) and their significant attributes.*

### 2.1 WORD2VEC

Word2Vec models work with an intuition as to train neural networks in order to predict a context given a word (Skip-Gram), or a word given a context (Continuous Bag of Words). *Figures 2.01 & 2.02* depict the architectures of the aforementioned models. Their implementation relies on the Feedforward Neural Net Language Model (introduced by Bengio et al.)

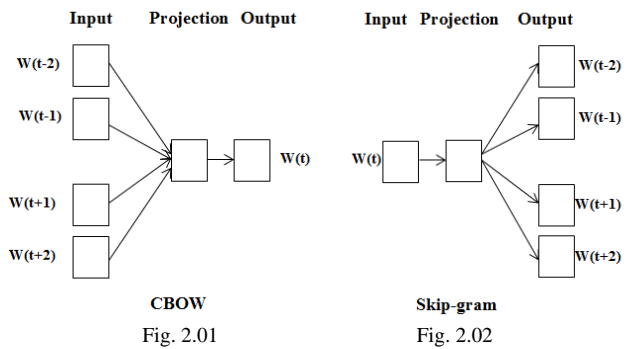

Fig. 2.01   Fig. 2.02

In the CBOW approach, the input layer takes the *N* previous words in *1xV* matrix, *V* being the size of the vocabulary. A projection using the shared projection matrix encoded by the projection layer is then made. A probability distribution over all the words in the vocabulary is computed with the help of a softmax activation function, the output layer shows probability of each word being the next one.

Skip-Gram approach works in the same way as CBOW, except that the task is to predict a context given a word. Technically speaking, the objective is to average log probability of the context of a given word. It is expressed in the following way:

$$J = \frac{1}{T} \sum_{t=1}^{T} \sum_{-c \leq j \leq c, j \neq 0} \log p(w_{t+j}|w_t)$$

Where *T* is the number of training words and *c* being the number of context words. *p(w<sub>t+j</sub>|w<sub>t</sub>)* is given by the softmax function expressed as the following:

$$p(w_o|w_i) = \frac{\exp(v'^\top_{w_o} v_{w_i})}{\sum_{w=1}^{W} \exp(v'^\top_{w} v_{w_i})}$$

Where *W* is the number of unique words in corpus $w_1 \ldots w_T$, and $v_w$ and $v_{w'}$ are the input and output vector representations of word *w*.

With an increase in the size of the vocabulary follows a significant increase in the computational cost of Word2Vec. Henceforth, one of the solutions often adopted is the usage of a hierarchical variant of the softmax function (Mnih et al., 2009). Being built upon a Huffman-tree base word frequency (most frequent words = shortest path from the root), *Hierarchical Softmax* reduces the computation complexity from *H* × *V* to $log_2(V)$ × *H*., where *H* is the size of the hidden layer and *V* being the vocabulary size.

Another solution backed by Mikolov et al. is a relatively novel technique called *Negative Sampling*. In negative sampling, the model updates the representations of a small number of words such that the network predicts an observed "positive" word pair (e.g., *Ice Cream*), and does not predict any of a number of "negative" pairs that are unlikely to be observed in the text (e.g. *ice future* or *ice happy*). Because a small number of negative samples are used—usually fewer than 20—a relatively small number of weights need to be updated at each iteration. Mikolov et al. further suggests sampling less on high frequency words for performance gains.

## 2.2 GloVe:

GloVe (Pennigton et al., 2014) relies on a co-occurrence matrix, *X*, with each cell, $X_{ij}$ containing the number of times a word, $w_i$ appeared in some context, $c_j$. By doing a simple normalization of the values for each row of the matrix, we can obtain the probability distribution of every context given a word. Consequently, the relation between two words can be calculated given by ratio between the probabilities of their context.

For words *i, j, k;* with $P_{ij}$ being the probability of *i, j* to be in the same context and $P_{ik}$ being the probability of *j, k* to be in the same context, the ration is expressed by the following:

$$F(w_i, w_j, \tilde{w}_k) = \frac{P_{ij}}{P_{ik}}$$

Following a series of assumptions and derivations, the authors finally arrive at simplified word vectors of the form:

$$\vec{w}_i^T \vec{w}_j + b_i + b_j = \log X_{ij}$$

Where $w_i$ is the representation of the i<sup>th</sup> word, w is the representation of the j<sup>th</sup> word, $b_i$ and $b_j$ are bias terms, and $\log X_{ij}$ is the co-occurrence count of words i and j. In contrast to word2vec, GloVe implements the following cost function to compute the best possible word representations of $w_i$, $w_j$:

$$J = \sum_{i,j=1}^{V} f(X_{ij})(w_i^T \widetilde{w}_j + b_i + \widetilde{b}_j - \log X_{ij})^2$$

*GloVe* further introduces a weighting scheme into the cost function of the model thereby avoiding log 0 errors and further reducing the negative effect of high frequency co-occurrences:

$$f(x) = \begin{cases} (x/x_{max})^\alpha & \text{if } x < x_{max} \\ 1 & \text{otherwise} \end{cases}$$

where *x* is the co-occurrence count, and *α* (an exponential weighting term of range: 0 and $X_{max}$) The performance of a GloVe model thus depends on the dimensionality of the word vector, $X_{max}$, $a$, and the size of the window used to compute co-occurrence statistics around each word.

Although it depends on the parameters of each model, GloVe is generally faster than Word2Vec (Pennigton et al., 2014) because it does not need to go through the entire corpus in each iteration. However, it is often reported by most papers that Word2Vec is slightly more accurate in terms of results of respective tasks (Berardi et al., 2015)

## 3. STABILITY CONSTRAINTS

The most fundamental cause for the inconsistency of contemporary word embeddings across various tasks is the lack of a universal applicability. One has to make necessary changes to the hyper-parameters (especially those concerning sub-sampling and context window size) depending on the context of its application and factors such as size of the corpus and the domain specificity of the data.

(Wendlandt et al., 2017) provides evidence of intrinsic factors that lead to instable embeddings in addition to initializing parameters.

Given a word $W$ and two embedding spaces $A$ and $B$, stability is defined as the percentage overlap between the ten nearest neighbors of $W$ in $A$ and $B$. The results of their preliminary analysis are depicted in *Figure 3.0*

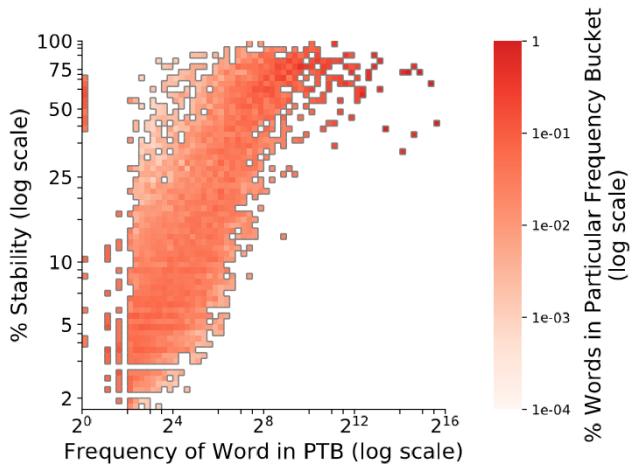

*Figure 3.0: Stability of word2vec as a property of frequency in the PennTreeBank (*Wendlandt et al., 2017)

From Fig 3.0, lower frequency words tend to have a lower stability and higher frequency words have a higher stability. However, the high variance in the graph indicates that the correlation is not linear. Wendlandt et al., use a ridge regression model to find additional factors that might cause this stability. Their training data includes features that take into account various word (eg: POS-tags, Wordnet Senses), data (eg: size of corpus, word frequency, domain), and algorithmic properties (eg: number of dimensions, GloVe/Word2Vec).

One of their prominent results is that there is more stability within models of the same domain as opposed to those of different domains. Insights form Wendlandt et al., could be exploited by building stable domain-specific models and aligning them together without compromising the stability. A promising approach to this is a technique known as *"retrofitting",* introduced by Faruqui et al. (2015) which combines embeddings learned from different models with a source of structured connections between words. The combined embedding achieves performance on word-similarity evaluations superior to either source individually.

With word embeddings being used as input resources in several downstream NLP tasks, it is necessary to ensure stability to make any qualitative judgment. Having said that, an absolutely stable embedding model would still fail to capture several features of language within its vector space.

At this point in the deconstructionist approach, I will make an assumption that their embedding models are stable and that their parameters are indeed optimized to their highest capacity. This will allow us to further question the composition of said embeddings using available literature and verify what aspects of language are still at stake with an NLP system that is built upon such vector-space models – if it is the case that they perform at their best baselines – and which of those shortcomings could theoretically be overcome.

## 4. PARALLELOGRAMIC CONSTRAINTS

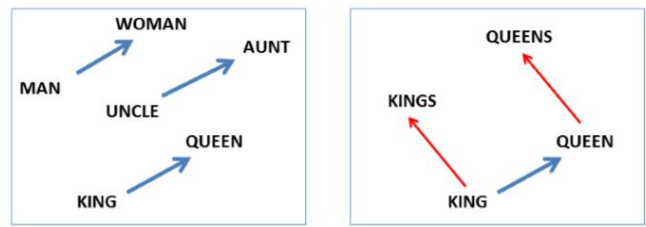

*Fig. 4.0 Linguistic relations modeled by linear vector offset* (Mikolov et al., 2013c)

Word Embeddings are popularized by their ability to draw analogical relations of type:

$V_{king} - V_{man} + V_{woman} = V_{queen}$

This method of deriving relationships between words from vector-offset has been referred to as the parallelogram model of analogy. Rogers et al., and Gladkova et al., refer to the aforementioned operation in particular as *'3CosAdd'*
In their analysis, they (Gladkova et al., 2016) show that what *3CosAdd* really captures is not analogy, but rather *relational similarity,* i.e. the idea that pairs of words may hold similar relations to those between other pairs of words. For instance, the relation between *cat* and *feline* is similar to the relation between *dog* and *canine*.

It is to be noted that this is *similarity* rather than identity*: "instances of a single relation may still have significant variability in how characteristic they are of that class"* (Jurgens et al., 2012) (Rogers et al., 2017)

*Analogy* as it is well known in domains of logic and philosophy follows the template:

> objects $X$ and $Y$ share properties $a$, $b$, and $c$;
> therefore, they may also share the property $d$.
> (3.1.1)

For example, both Earth and Mars orbit the Sun, have at least one moon, revolve on axis, and are subject to gravity; therefore, if Earth supports life, so could Mars (Bartha, 2016).

**Why is it then that a majority of research from the domain of distributional semantics refer *to 3CosAdd* as an evaluation metric for analogical reasoning and not relational similarity?**

It seems to be the case that Turney distinguishes between attributional similarity between two words and relational similarity between two pairs of words. On this interpretation, *two word pairs that have a high degree of relational similarity are analogous* (Turney, 2006).

It is this use of the term "analogy" that Mikolov et al. (2013c) followed in proposing the 3*CosAdd* method which later came to be standardized as a baseline evaluation metric for all those in the scientific community who began further investigating embeddings.

Therefore, it is to be noted that the *parallelogram model of analogy* (3CosAdd), in its core is simply an estimation of relational similarity, and could be formulated in the following way:

> Given a pair of words $a$ and $a_0$,
> find how they are related and then find word $b_0$,
> such that it has a similar relation with the word $b$.

One could contrast the above template with that of an analogical relation (in the template 3.1.1) and observe the extent of its separation from the actual structure.

(Levy and Goldberg, 2014) further support this inefficiency of the analogy task in that the accuracy varies widely between categories. This discontinuity between actual analogy and that claimed by *3CosAdd* is further backed by Gladkova's results in that the Glove-Model that has an accuracy above 80% (in terms of 3CosAdd) obtains a mere 28% accuracy on the BATS (Bigger Analogy Test Set) (Gladkova et al., 2016).

**What exactly does the parallelogram model of analogy do?**

Mikolov et al. (2013c) in formulating 3CosAdd excludes the three source vectors $a$, $a_0$ and $b$ from the collection possible answers.

For Example, in answering statement:

$$V_{king} - V_{man} + V_{woman} = \_ ?$$

the model ignores $V_{king}$, $V_{man}$, and $V_{woman}$ from the set of possible answers.

Although this may sound to be a valid exclusion to have, analyzing the inclusion of the source vectors can give use insights on what *3CosAdd* actually computes. Linzen (2016) showed that if it is not for the exclusion, the accuracy drops dramatically, hitting zero for 9 out of 15 Google test categories.

Wendlandt et al., further investigate this property on how different categories of data perform if the source vectors are included. The rows of Fig. 2 represent all questions of a given category, with darker color indicating higher percentage of predicted vectors being the closest to $a$, $a_0$, $b$, $b_0$, or any other vector.

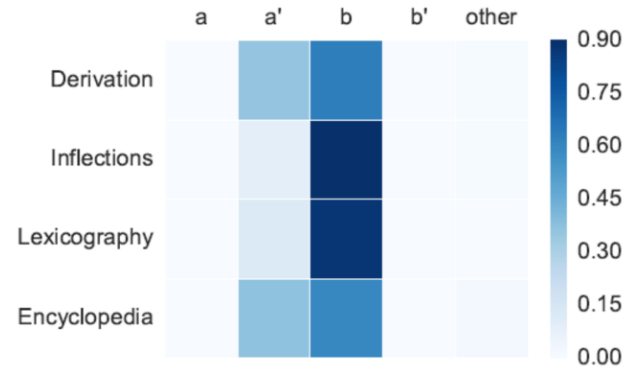

*Figure 2: The result of a - a' + b calculation on BATS. Source vectors a, a' and b are not excluded (*Wendlandt et al., 2016)

Fig. 2 shows that if we do not exclude the source vectors, $b$ is the most likely to be predicted; in derivational and encyclopedic categories $a_0$ is also possible in under 30% of cases. $b_0$ is as unlikely to be predicted as $a$, or any other vector. This experiment suggests that the addition of the offset between $a$ and $a_0$ typically has a very small effect on the $b$ vector – not sufficient to induce a shift to a different vector on its own. This would in effect limit the search space of *3CosAdd* to the close neighborhood of the $b$ vector.

Consequently, targets that are further away in the vector space have worse chances of being recovered. With their being a huge number of linguistic relations, many of which are context-dependent, no model could possibly hold all related word close in the vector space.

Linzen (2016) further points out that in the plural noun category of the Google test set, 70% accuracy was achieved by simply taking the closest neighbor of the vector b, while 3CosAdd improved the accuracy by only 10%. This is expected because the singular and plural forms of a particular word were similar to an extent that subtracting them is nearly a null vector.

For instance,

$$V_{dogs} - V_{dog} + V_{cats} = V_{cats} \; ; \; if \; V_{dog} \approx V_{dogs}$$

But since $V_{cats}$, being one of the source vectors, is excluded, *3CosAdd* would output $V_{cat}$ as it is the closest neighbor to $V_{cats}$.

In section 2, we have made the assumption for hyper-parameters used in a model to be the most optimal for a given application. Similarly, considering the inadequacy of parallelogram model of analogy, the assumption we make at this stage is that there exists an alternative method that exploit vector-offsets (distances) in the most efficient way.

Gladkova et al. (2016) in fact provides such alternative methods (3CosAvg and LRCos) that perform significantly better in relation to the much criticized 3CosAdd. However, the assumption we make is not necessarily of an already existing method. The objective here is to consider what other linguistic features fail to be well accounted for provided there is a theoretically perfect method for exploiting spatial distances in vector space models.

## 5. GEOMETRIC CONSTRAINTS

Distance metrics in vector spaces must obey certain geometric constraints, such as *symmetry* (the distance *from x to y* is the same as the distance from *y to x*) and *the triangle inequality* (if the distance between *x and y* is small and the distance between *y and z* is small, then the distance between x and z cannot be very large) (Griffith et al., 2017). If we take it for granted that relations between words can be reflected by distances in a vector space, it would imply that representations of these words are also subjected to the same geometric constraints.

However, research in cognitive science has criticized this property of spatial representations of linguistic similarity because aspects of human semantic processing do not conform to these constraints (Tversky, 1977). For instance, people's interpretation of semantic similarity do not always obey the *triangle inequality*:

Words $w_1$ and $w_3$ are not necessarily similar when both pairs of $(w_1,w_2)$ and $(w_2,w_3)$ are similar.

While *"asteroid"* is very similar to *"belt"* and *"belt"* is very similar to *"buckle"*, *"asteroid"* and *"buckle"* are not similar (Griffith et al., 2007). Another famous example that violates the principle of symmetry is that people judge *North Korea* to be more similar to *China* than the other way around.

The inconsistency between human judgements of similarity and the relational similarity in spatial representation is further verified by Griffiths et al. (2017) by a series of experiments.

In order to check the prevalence of asymmetry in human judgements, they isolated certain word pairs that have multiple relations and share one of their less salient relations with another word pair. For example, when presented with *angry : smile – exhausted : run*, one might think, *"an angry person doesn't want to smile" and "an exhausted person doesn't want to run,"* but when presented with *exhausted : run – angry : smile*, one might think, *"running makes a person exhausted, but smiling doesn't make a person angry."* (Griffith et al., 2017).

They've evaluated these word pairs on 1,102 participants from Amazon Mechanical Turk. The previously mentioned example of *angry: smile and exhausted : run* elicited higher ratings in the direction shown here (4.76 mean rating) than in the opposite direction (2.36 mean rating).

As another example, people rated *hairdresser : comb – pitcher : baseball* as more relationally similar (6.10 mean rating) than *pitcher : baseball – hairdresser : comb* (4.84 mean rating). In the first presentation order, participants might be thinking that *"a hairdresser handles a comb"* and *"a pitcher handles a baseball,"* whereas in the second presentation order, they might be thinking *"a pitcher plays a specific role in baseball,"* which doesn't fit with *hairdresser : comb* (Griffith et al., 2017) .

A similar experiment was done to verify the violation of Triangle Inequality in human judgements. Triads of words pairs were selected for which the authors expected people's relational similarity judgments to violate the triangle inequality, such *as nurse : patient, mother : baby,* and *frog : tadpole.*

This triad violates the triangle inequality because *nurse : patient :: mother : baby* is a good analogy (relationally similar), and so is mother : baby :: frog : tadpole, but *nurse : patient :: frog : tadpole* is not. In this example, the middle pair has multiple relations and shares one of them with the first pair and a different one with the last pair. The participants were presented two word pairs in each of the triads and were asked to rate the quality of the analogy rather than relational similarity. The last pair was shown to have been allocated a higher score by embedding models while the participants were clearly able to articulate the distinction.

The results of their experiment are shown below:

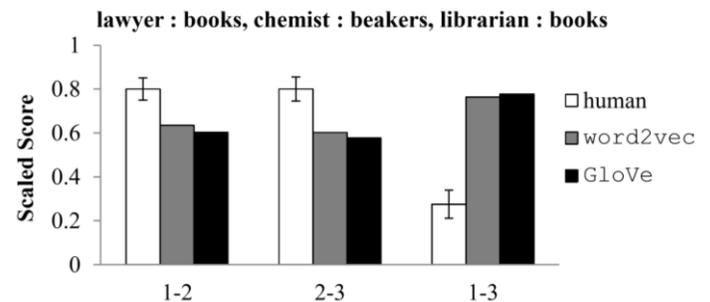

*Fig 5.0: Mean human ratings and predicted relational similarities.* [1,2,3] refer to triads of type [lawyer : books, chemist : beakers, librarian : books].

Consequently, [1-2] → [lawyer : books :: chemist : beakers] and [1-3] → [lawyer : books :: librarian : books].

The greater repercussion of the incompatibility of spatial geometry to account for the intricacies of human judgement is that only a small subset of linguistic properties will be taken into account. Higher level properties such as Synonymy-Antonymy, Hypernymy-Hyponymy rely on the asymmetrical nature of language and will fail to have traces in the vector representations of words. [1]

Therefore, even if we ignore the constraints of instability and parallelogramic constraints, we still fail to shape a universal embedding that could capture the essence of language owing to the geometric constraints of vector space models. However, recent developments that propose modifications to Mikolov's model to take into account some of these linguistic properties that geometric models natively fail to. These developments are explored in the following section:

### 5.1 SYNONYMY-ANTONYMY

Solutions proposed to account for synonymy-antonymy rely on contextual patterns such as "*x and not y*" or "from *x to y*", or an external thesaurus. Although they perform well on certain cases, their usage is limited to the extents of the pattern. Scheible et al. points out that the occurrence of these pre-defined contextual patterns is not guaranteed and that it could further hamper the normal performance of the model. However, a German group of researches, demonstrated relative success using exclusively a vector space model without relying on any external resources or contextual patterns. (Scheible et al., 2017)

Their hypothesis was that not all parts of speech are useful for distinguishing antonyms from synonyms. They proposed the usage of multiple vector spaces, each taking its shape from co-occurrence matrices of different POS-tags.

Their results show that verbs serve as the most distinguishing features to identify synonymy-antonym. Their model, consisting of a Decision Tree classifier based on cosine difference values, obtained an accuracy of 70.6%. Although this is the highest efficiency so far in this domain, it is not optimal enough to be used for a clear distinction.

Moreover, the experiment was limited to a German corpus and although the authors claim that the same can be applied to other languages, no such implementations have been put forth.

### 5.2 POINCARE EMBEDDINGS FOR HYPERNYMY AND HYPONYMY:

Poincaré embeddings were proposed (Nickel and Kiela 2017) with an intuition that Euclidean vector spaces do not account for the property of latent hierarchical structure of language owing to their incompatibility with asymmetry. (a → b should be different from b → a for hierarchy to exist).

As opposed to Euclidean vector spaces, Poincare employs a hyperbolic space, an n-dimensional Poincare ball, to capture this hierarchy without compromising the similarity measures that embeddings otherwise capture anyway.

The learnt embeddings capture these notions of both similarity and hierarchy in the following ways: by placing connected nodes close to each other and unconnected nodes far from each other, and further computing the distance between these nodes serves as a measure for similarity; hierarchy can be captured by placing nodes lower in the hierarchy farther from the origin, and nodes higher in the hierarchy close to the origin (center of the Poincare ball).

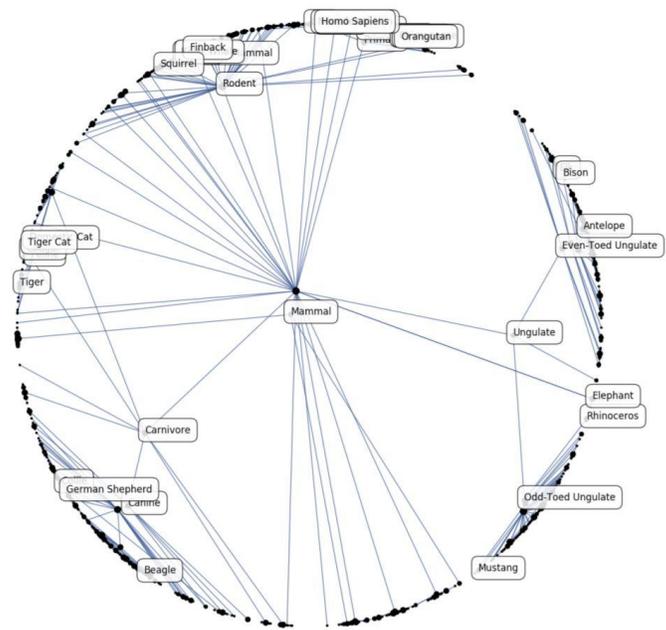

*Fig. 5.2.1:* Learnt Embeddings presented in *Nickel and Kiela (2017)*

Although the notion of using a non-euclidean vector space seems motivating, developments in Poincare Embeddings are limited to supervised methods in that they leverage external information from knowledge bases such as WordNet in addition to raw text corpora.

However, Tifrea et al. proposes to adapt the GloVe algorithm to hyperbolic spaces and to leverage a connection between statistical manifolds of Gaussian distributions and

---

[1] Note to Advisor: From our discussion on asymmetry, you have proposed add two relationships such that A→B is different from B→A. But the dimensional explosion of this approach sets a high computational cost (to avoid which are word embeddings initiated in the first place)

hyperbolic geometry thereby making the model entirely unsupervised.

It is at this stage of the deconstruction approach that I shall make a considerably significant assumption. The assumption is that there exists some method that efficiently encodes properties of asymmetry and triangle inequality. This task is not extremely intensive as developments in Poincare embeddings already show progress.

## 6. ETHICAL CONSTRAINTS

With an objective to accommodate language and its contextual, cultural, and structuralist essence, the embedding models in discussion are trained on natural user-generated data. However, the use of unsupervised algorithms trained on user-generated data poses the risk of reproducing the bias present in the data. Female/male gender stereotypes have appeared on word embeddings trained on Google News data (Bolukbasi et al., 2016). The obtained word embeddings connect *'queen'* to *'woman'* the same way they connect '*receptionist'* to *'woman'*.

The existence of genderless nouns in English can be used to analyse stereotypes by looking at the associations between those nouns and the words he and she. For example, the following equality has been observed (Bolukbasi et al., 2016) in word vectors trained by GloVe:

$$v_{(man)} - v_{(woman)} \approx v_{(programmer)} - v_{(homemaker)}$$

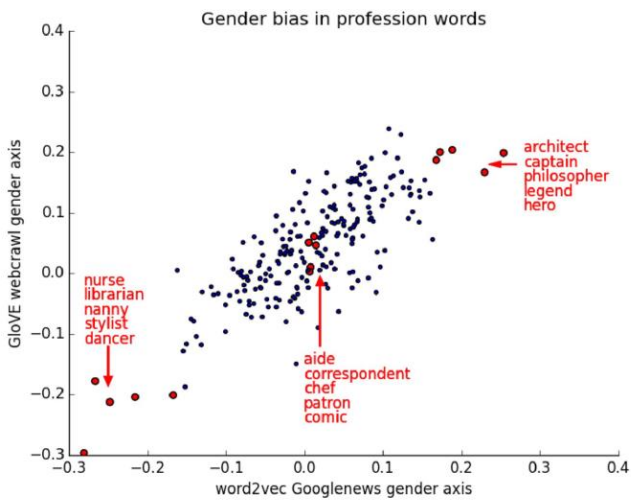

*Fig 6. Gender Bias in Word2Vec and GloVe models train on GoogleNews Dataset*

The combination of large amounts of decentralized, user-generated data with unsupervised algorithms that find hidden patterns in that data can lead to biased results in two ways.

In the first instance, it can reproduce biases contained in the original data. In the second instance, the algorithms can enhance previous biases by identifying biased parameters as fundamental characteristics of the concept. The bias in the training data is therefore amplified as embedding algorithms are designed to generalize models using the information contained in the original data. This can lead to higher weights to be allocated to biased parameters.

The use of biased word embeddings in different applications pose a threat to fair decision making processes as the inherent bias is automatically passed on to any application that uses the word embeddings, perpetuating in this way cultural stereotypes. Furthermore, the results of word embedding algorithms affect how we relate to the world in web search results, language generation applications such as customer service or social media bots and news summaries generation.

## 7. DISTRIBUTIONAL CONSTRAINTS

As discussed in the previous section, embedding models being distributional models rely on corpus data for learning concepts of properties. The Distributional Hypothesis states that the meaning of words can be inferred from their linguistic environment (Harris, 1954). This hypothesis lies at the heart of all embedding models and is the presumption by which they approximate the meaning of words by considering the statistics of their co-occurrence with other words in the lexicon.

Most of the constraints on embedding models that have been discussed in the preceding sections could be attributed to a broader constraint put forth by the aforementioned hypothesis. It has been further argued, in recent years, that distributional models can never absolutely grasp the meaning of words, as many aspects of language are not necessarily present in language but outside it (in that some aspects of semantics can have grounding in the physical world) (Andrews et al., 2009).

Rubinstein et al., (2015) verifies the claims against the distributional hypothesis using a comprehensive evaluation. They split meanings of words into two categories: taxonomical and attributive.

Properties that make the compositional conception of terms are classified as taxonomical (eg: *Apple is a fruit*) while properties that contribute to adding detail to terms are classified as attributive (eg: *an apple is red, an apple is round*). The results of their analysis show that in a binary property prediction task (Attributive evaluation: *is an apple red?*; Taxonomical evaluation*: is apple a fruit?),* attributive properties showed an F-score of *0.37* in comparison to a 0.74 on taxonomical properties.

This suggests that the distributional hypothesis may not be equally applicable to all types of semantic information, and in particular, it may be limited with respect to attributive properties. Therefore, abstract concepts such as *'dangerous'*, *'cute'* or perceptual properties such as color will remain to be poorly represented as long as one relies purely on the corpus data.

Therefore, a multi-modal approach that does not purely rely on corpus data could potentially lift the constraints set forth by the distributional hypothesis and in doing so, might show progress in terms of pushing the constraints that have been discussed above.

There have been developments that show promise in a multi-model approach: Collell and Moens (2016) show how representations from computer vision models can help improve these predictions. Barsalou (1999) claims that many human modalities such as conceptual/perceptual systems cooperate with each other in a complex way and inence word meanings.

Park and Myaeng (2017) build on top of the aforementioned claims in order to provide evidence on such studies on the multifaceted nature of word meanings. They propose a polymodal-word embedding that takes into account the following types of resources as inputs to individual embeddings:

- **Emotion** (employs the NRC Emotion Lexicon to reflect the emotional characteristics of words);
- **Perception** (Vision specific word embeddings by trainings images and sentences together);
- **Cognition** (retrofitted embeddings from WordNet ontology containing relations of synonymy, hypernymy, and homonymy);
- **Sentiment** (employs SentiWordNet3.0 for automatic annotation interms of the degree of positivity/negativity);
- **Linear context** (standard word2vec embedding with skip-gram and negative sampling);
- **Syntactic context** (a skip-gram model with context-window based on a dependency parse tree).

These individual embeddings are amalgamated into an ensemble following the methods prescribed Faruqui et al. which were discussed in *Section 3*. Their comparative study with word2vec and Glove indicates that the proposed embedding performs better in similarity and hypernym prediction tasks.

The model also depicts a higher accuracy in concepts that are associated with perceptual and cognitive capacities. However, since the method largely relies on labeled data, there is a limitation in terms of scalability.

# 8. CONCLUSION

Through a series of deconstructions, this paper brings to surface the shortcomings of word embeddings at various level and the degree by which these shortcomings can be solved (if they can be solved). Another interesting insight is that word embedding models may miss out on some fundamental elements of semantics as long as they rely purely on corpus data.

We observe traces of developments in contemporary embedding models in reaching towards the full capacities of the "*Derridian Embedding Framework*" that was introduced in Section 1. However, the temporal component of Derrida's *Differance* remains to be an unexplored avenue with recent developments only accounting for the spatial component.

With most research building on top of already existing models / methods, there is no incentive to revert back to the pre-Mikolovian times of word embedding (why start over?). Therefore, there is not enough research in the domain of distributional semantics that proposes an alternative approach to embedding models that account for temporality.

With lessons learnt from the shortcomings and successes of vector space models, a new embedding model that takes into account both the spatiality and the temporality of language will probably induce a paradigm shift, and possibly reach the capacities of a Derridian Embedding Model.